\documentclass[journal]{IEEEtran}

\usepackage{amsmath,amsfonts,amssymb}
\usepackage{color}
\usepackage{cite}
\usepackage{mathtools,trimclip,stackengine,scalerel}
\usepackage{amsfonts}
\usepackage{hyperref}
\usepackage{xcolor} 
\usepackage{cases}
\usepackage{multirow}
\usepackage{booktabs}
\usepackage[caption=false]{subfig}
\usepackage{diagbox}
\usepackage{nicematrix}
\usepackage[bottom,symbol]{footmisc}
\usepackage[T1]{fontenc}
\usepackage[utf8]{inputenc}
\usepackage{babel}
\usepackage{tabularx}
\usepackage[export]{adjustbox}
\usepackage{booktabs}
\usepackage{varwidth}
\usepackage[ruled, lined, linesnumbered, commentsnumbered, longend]{algorithm2e}
\usepackage{algpseudocode}
\usepackage{mathtools}
\usepackage{commath}
\usepackage{enumitem}
\usepackage{float}
\usepackage{ntheorem}
\usepackage{soul}
\usepackage{graphicx}
\usepackage{makecell}

\definecolor{m}{RGB}{240,0,240}

\newcommand*\xbar[1]{%
  \hbox{%
    \vbox{%
      \hrule height 0.5pt 
      \kern0.5ex
      \hbox{%
        \kern-0.1em
        \ensuremath{#1}%
        \kern-0.1em
      }%
    }%
  }%
}

\SetCommentSty{mycommfont}

\setenumerate[1]{label={(\arabic*)}} 
\newsavebox\tmpbox
\newcommand\hvec[1]{\ThisStyle{%
  \setbox0=\hbox{$\SavedStyle#1$}
  \setbox2=\hbox{$%
    \clipbox{0pt{} \dimexpr\ht0+1.68\LMpt{} -.2\LMpt{} 0pt}{%
      $\SavedStyle\mathaccent"017E{\phantom{\SavedStyle #1}}$}\kern-.2\LMpt$}
    \ensurestackMath{\stackengine{1.3\LMpt}{\SavedStyle#1}{\copy2}{O}{c}{F}{F}{S}}
}}
\newcommand{\algrule}[1][.2pt]{\par\vskip.5\baselineskip\hrule height #1\par\vskip.5\baselineskip}

\ifCLASSINFOpdf
\else
\fi

\def\*#1{\boldsymbol{#1}}

\newcommand{\VW}{\mathbf{W}}
\newcommand{\VTheta}{\mathbf{\*\Theta}}

\newcommand{\appropto}{\mathrel{\vcenter{
  \offinterlineskip\halign{\hfil$##$\cr
    \propto\cr\noalign{\kern2pt}\sim\cr\noalign{\kern-2pt}}}}}

\begin{document}

\title{Training of Spiking Neural Networks with Expectation-Propagation}

\author{Dan Yao,
        Steve McLaughlin,~\IEEEmembership{Fellow,~IEEE,}
        and~Yoann Altmann,~\IEEEmembership{Member,~IEEE}
\thanks{D. Yao, S. McLaughlin and Y. Altmann are with the School of Engineering and Physical Sciences, Heriot-Watt University, EH14 4AS, Edinburgh, United Kingdom, e-mail: Y.Altmann@hw.ac.uk.}
\thanks{This work was supported by the UK Royal Academy of Engineering under the Research Fellowship Scheme (RF201617/16/31) and by the Engineering and Physical Sciences Research Council of the UK (EPSRC) Grant numbers EP/T00097X/1 and EP/S000631/1 and the UK MOD University Defence Research Collaboration (UDRC) in Signal Processing.}}


\maketitle

\begin{abstract}
In this paper, we propose a unifying message-passing framework for training spiking neural networks (SNNs) using Expectation-Propagation. Our gradient-free method is capable of learning the marginal distributions of network parameters and simultaneously marginalizes nuisance parameters, such as the outputs of hidden layers. This framework allows for the first time, training of discrete and continuous weights, for deterministic and stochastic spiking networks, using batches of training samples. Although its convergence is not ensured, the algorithm converges in practice faster than gradient-based methods, without requiring a large number of passes through the training data. The classification and regression results presented pave the way for new efficient training methods for deep Bayesian networks.  

 
\end{abstract}

\begin{IEEEkeywords}
Spiking neural networks, Expectation-Propagation, 
deep Bayesian networks, uncertainty quantification.
\end{IEEEkeywords}

\IEEEpeerreviewmaketitle

\section{Introduction}
\label{sec: Introduction}
\IEEEPARstart{S}{piking} neural networks (SNNs), as the third generation of neural network models \cite{maass1997networks}, have emerged as a powerful approach to information processing inspired by biological brains \cite{tavanaei2019deep, taherkhani2020review}. Unlike conventional artificial neural networks (ANNs)
that transmit information using continuous values, SNNs operate on binary spike trains. By employing spiking neurons as the computational units, temporal information is encoded in the dynamics of the spikes, which can mimic the behavior of real biological neurons in response to external stimuli. Such characteristics make SNNs particularly suited to perform energy efficient computation\cite{indiveri2015memory, tavanaei2019deep, kasabov2019time}. 
In recent years, SNNs have demonstrated promising potential in a variety of applications where higher energy efficiency and lower computational cost is important, such as image classification \cite{mozafari2019bio}, spiked-based object detection \cite{kim2020spiking}, robot control \cite{bouganis2010training} and 
 neuromorphic simultaneous localization and
mapping (SLAM) \cite{kreiser2018pose, weikersdorfer2013simultaneous}. 

However, SNN training remains challenging and has impeded their more widespread deployment. One of the main difficulties arises from the non-differentiability of the typical spiking activation. To address this challenge, a variety of SNN training methods have been proposed and they can be categorized into two main types. Indirect training methods typically train traditional (non-spiking) ANNs and convert the architectures into SNNs \cite{rueckauer2017conversion, o2013real, diehl2015fast, sengupta2019going}. 
Conversely, most direct training methods employ classical surrogate gradient methods \cite{neftci2019surrogate, zenke2021remarkable}, leaving the original training loss unchanged. However, it is possible to reformulate the training task, using probabilistic models and methods \cite{jang2019introduction, brea2013matching, maass2014noise}, to derive new objective functions that are either differentiable, or do not rely on gradient-based optimization. Moreover, Bayesian methods can also potentially provide uncertainty quantification tools such as parameter posterior distributions that can be particularly useful to assess relevance of features \cite{gast2018lightweight, skatchkovsky2022bayesian, kendall2017uncertainties, gawlikowski2023survey}. The method proposed in this paper belongs to this last family of direct training methods. 

SNN architectures can also be divided in two families based on their neuron types. The most popular SNNs for computer vision tasks are networks whose activation functions are deterministic, e.g. the Heaviside function, with parameters/weights treated as deterministic variables. We refer to those networks as deterministic SNNs. Stochastic SNNs on the other hand, rely on recent studies in cognitive science and neuroscience that have shown that neurons and synapses are inherently stochastic \cite{deco2009stochastic, buesing2011neural}. In this work, we propose a framework able to handle deterministic and stochastic SNN training using Bayesian inference, in a similar fashion to \cite{jang2021bisnn}. This allows prior knowledge such as pre-training results or weight constraints to be incorporated in a principled manner.

As for most deep-learning architectures, exact Bayesian computation is not tractable given the large number of network unknowns and in this work, we propose a flexible network based on Expectation-Propagation  (EP) \cite{MinkaEP, wainwright2008graphical}. EP-based methods are a family of variational inference tools which allow for message-passing type of updates. In contrast to most variational Bayes methods which also rely on Kullback-Leibler divergences as similarity metrics between distributions, EP does not require distributions to be differentiable.
From a methodological viewpoint, EP-based methods have been recently proposed to train deep Bayesian networks \cite{jylanki2014expectation, ghosh2016assumed}, including spiking networks \cite{shen2023probabilistic} and networks with binary/discrete weights \cite{soudry2014expectation}. However, methods such as probabilistic backpropagation \cite{hernandez2015probabilistic} are not explicitly designed to allow multiple passes (epochs) through the data or process simultaneously multiple training samples, as intended here. Thus, they tend to provide unreliable uncertainty estimates. We combine EP and stochastic EP (SEP) \cite{li2015stochastic, dehaene2018expectation} in a modular fashion and the method, tailored here to train SNNs, can be generalized to other, potentially recurrent, neural networks. The resulting algorithm still preserves the notions of forward and backward passes through the network, as in probabilistic backpropagation \cite{hernandez2015probabilistic}. 

The main contributions of the paper can be summarized as follows.
\begin{itemize}
    \item We derive an EP-based method able to train feedforward neural networks using mini-batches. This method, applied to spiking architectures without recurrent neurons, serves as a building block to train recurrent SNNs. 

    \item The method is able to handle, in a unified framework, deterministic and stochastic SNNs as well as the estimation of discrete and continuous weights.  

    \item In addition to point estimates, the method provides marginal weight uncertainties and allows the computation of predictive distributions at inference time without resorting to Monte Carlo sampling (of the network weights or stochastic activation functions). 
\end{itemize}

The remainder of the paper is structured as follows. Section \ref{sec: Preliminaries} introduces the neuron model and SNN architectures considered. Section \ref{sec: Bayesian model} derives the likelihood associated with SNNs and the resulting posterior distribution of the network parameters. Section \ref{sec: EP_BNN} details how to compute approximating distributions by combining stochastic/average EP updates. Section \ref{sec: Algorithmic considerations}
discusses convergence and the use of large datasets. 
The potential of the method is illustrated using results obtained for classification and regression problems in Section \ref{sec: Experiments}. Conclusions and potential future work are finally reported in Section \ref{sec: Conclusions and Future Work}.

\section{Preliminaries}
\label{sec: Preliminaries}
We consider a general supervised training task where a set of i.i.d. input/output data pairs  $\mathcal{D}_N =\{ (\*x_{n}, \*y_{n})\}_{n=1}^N$ is available. In the context of SNNs, the input $\*x_{n}$ is typically a binary set of inputs with multiple timestamps, but in this paper we focus on single timestamps. In contrast, $\*y_{n}$ corresponds to the network output, which can be binary values (spiking output, e.g. for classification problems), real values (e.g. for regression problems), or a combination of those, depending on the activation function used in the output layer (see Section \ref{subsec: Spiking neurons under Spike Response Mode} below). 

\subsection{Spiking neurons under Spike Response Model}
\label{subsec: Spiking neurons under Spike Response Mode}

In this work, we consider a simplified Spike Response Model (SRM) \cite{gerstner1995time} for our spiking neurons. Following SRM, the spiking activity of  neuron $i$ is determined by a scalar internal state, known as membrane potential $u_{i,n}$, which can be expressed as
\begin{equation}
u_{i,n} =  \sum\limits_{j \in \mathcal{P}_i} w_{ij}v_{j,n}, 
\label{Eq: membrane potential_u}     
 \end{equation}
where  $w_{ij}$ is the synaptic weight connecting neuron $i$ to its presynaptic neuron $j$ and $\mathcal{P}_i = \{j | j  \text{ presynaptic to } i\}$. Moreover, $v_{j,n}$  represents the \textit{feedforward} (spiking) signal from the neuron $j$ of the previous layer, for sample $n$. Note that to simplify the algorithm description, Eq. \eqref{Eq: membrane potential_u} does not include biases, but these can easily be included. 

The spiking output of each neuron $v_{i,n}$ is modeled using a stochastic, potentially degenerate distribution $f(v_{i,n}|u_{i,n})$, conditioned on the neuron membrane potential $u_{i,n}$. 
The typical options in this paper for  $f(v_{i,n}|u_{i,n})$ are
\begin{eqnarray}
    \left\{ \begin{array}{cc}
	f(v_{i,n}|u_{i,n})  = &    \left\{ \begin{array}{c}
    \delta(v_{i,n}-1) ~~ \textrm{if} ~~u_{i,n}\geq 0\\
    \delta(v_{i,n}) ~~ \textrm{if} ~~ u_{i,n}<0
    \end{array} \right. ,\\
	f(v_{i,n}|u_{i,n})  = &  \mathcal{B}ern\left(\sigma(u_{i,n})\right)\\
\end{array} \right. ,
\label{Eq: spiking emission function}    
\end{eqnarray}
where $\delta(\cdot)$ is the Dirac delta function and $\sigma(u_{i,n}) = (1+\text{exp}^{-u_{i,n}})^{-1}$ is the usual sigmoid function. These options correspond to classical deterministic SNNs (with Heaviside activation) and probabilistic SNNs \cite{jang2019introduction}, respectively.  However, other activation functions can also be used. For instance, if the output of the network is continuous, $f(v_{i,n}|u_{i,n})$ can be set as a Gaussian distribution. 

\subsection{Network architecture and training task}
\label{subsec: Network architecture and training task}


In this work, we consider layered network architectures. More precisely, the feedforward networks consist of $L$ fully connected layers. 
Each layer $\ell$ has $V_\ell$ spiking neurons, for $\ell=1,\dots, L$. At the $\ell$-th layer,  we denote by $\VW^{(\ell)} \in \mathbb{R}^{V_\ell \times V_{\ell-1}} $ the weight matrix,  which leads to 
\begin{eqnarray}
\label{eq:linear_layer}
    \*u_{n}^{(\ell)} & = & \VW^{(\ell)} {\*v}_{n}^{(\ell-1)} \in \mathbb{R}^{V_\ell \times 1},
\end{eqnarray}
 where $ {\*v}_{n}^{(\ell-1)}$ denotes the forward signal (or output) of layer $(\ell-1)$ for samples $n$ and where the same convention is used to define $\*u_{n}^{(\ell)}$ (with $\*v_{n}^{(0)}=\*x_n$). 

 The main training task addressed in this paper consists of estimating (the distribution of) the network trainable parameters $\mathcal{W} = \{\VW^{(\ell)}\}_{\ell=1}^L$. However, since the mappings $f(\*v_{n}^{(\ell)}|\VW^{(\ell)},{\*v}_{n}^{(\ell-1)})$ can be stochastic, the (spiking) output of each layer is a random vector, which propagates uncertainties in the subsequent layers of the network. Thus, the estimation of $\mathcal{W}$ also requires the estimation or marginalization of a range of nuisance parameters, as will be discussed in the next section.

\section{Bayesian model for SNN training}
\label{sec: Bayesian model}
\subsection{Forward model for a single training sample}
The layered structure presented in the previous section is particularly well suited to derive a statistical model relating the network inputs to its outputs.

Using Eqs. \eqref{Eq: spiking emission function} and \eqref{eq:linear_layer}, the stochastic model linking the input ${\*v}_{n}^{(\ell-1)}$ to the output ${\*v}_{n}^{(\ell)}$ of layer $\ell$ can be expressed as $f(\*v_{n}^{(\ell)}|\VW^{(\ell)}, {\*v}_{n}^{(\ell-1)})$.
However, to facilitate inference using message passing, we extend this model, using $\*u_{n}^{(\ell)}$ as auxiliary variable, such that\\
$f(\*v_{n}^{(\ell)},\*u_{n}^{(\ell)}|\VW^{(\ell)}, {\*v}_{n}^{(\ell-1)})=$
\begin{eqnarray}f(\*v_{n}^{(\ell)}|\*u_{n}^{(\ell)})f(\*u_{n}^{(\ell)}|\VW^{(\ell)}, {\*v}_{n}^{(\ell-1)}),
\end{eqnarray}
with 
\begin{eqnarray}
f(\*u_{n}^{(\ell)}|\VW^{(\ell)},{\*v}_{n}^{(\ell-1)})=\delta\left(\*u_{n}^{(\ell)}-\VW^{(\ell)}\*v_{n}^{(\ell-1)}\right).\end{eqnarray}
Let $\*\theta_{n} = \{ \{\*u_{n}^{(\ell)}\}_{\ell=1}^L, \{\*v_{n}^{(\ell)}\}_{\ell=1}^{L-1}\}$ gather all the unknown latent variables associated with the $n$-th training sample $\*x_n$. When feeding the network with $\*v_{n}^{(0)} = \*x_{n}$ and $\mathcal{W} = \{\VW^{(\ell)}\}_{l=1}^L$, we obtain
\begin{eqnarray}
\label{eq:forward_model_one_sample}
f(\*\theta_{n},\*v_{n}^{(L)}|\mathcal{W},\*v_{n}^{(0)})=\prod_{\ell=1}^L f(\*v_{n}^{(\ell)},\*u_{n}^{(\ell)}|\VW^{(\ell)},{\*v}_{n}^{(\ell-1)}).
\end{eqnarray}

\subsection{Forward model for the whole training set} 
A few additional notations are needed to simplify the presentation of the joint forward model, namely  $\VTheta=\{\*\theta_{n}\}_n$, as well as $\*U^{(\ell)}=\{\*u_{n}^{(\ell)}\}_{n}$, $\*V^{(\ell)}=\{\*v_{n}^{(\ell)}\}_{n}$, $\*U=\{\*U^{(\ell)}\}_{\ell=1}^L$ and $\*V=\{\*V^{(\ell)}\}_{\ell=1}^{L-1}$, such that $\VTheta=\{\*V,\*U\}$. Note that the outputs of the last layer are not included in $\*V$, since they are known/fixed during training, and are thus not treated as latent variables.
The network weights $\mathcal{W}$ are global parameters shared by all the training samples. Assuming the training data are mutually independent, Eq. \eqref{eq:forward_model_one_sample} leads to 
\begin{eqnarray}
\label{eq:forward_model_all_samples}
f(\VTheta,\*V^{(L)}|\mathcal{W},\*V^{(0)})=\prod_{n=1}^N f(\*\theta_{n},\*v_{n}^{(L)}|\mathcal{W},\*v_{n}^{(0)})
\end{eqnarray}

\subsection{Bayesian training of SNNs}
Adopting a Bayesian approach, $\mathcal{W}$ can be assigned a prior $p_0(\mathcal{W})$ and one can conceptually obtain the marginal posterior distribution of $\mathcal{W}$ given the training dataset $\mathcal{D}_N$, 
\begin{equation}
\label{eq:marginal_posterior}
p\left(\mathcal{W}|\mathcal{D}_N\right) \propto p_0\left(\mathcal{W}\right)\left[\displaystyle \int f(\VTheta,\*V^{(L)}|\mathcal{W},\*V^{(0)}) d \VTheta\right], 
\end{equation}
where the local/nuisance latent parameters in $\VTheta$ are marginalised. However, the high dimensional integral in Eq. \eqref{eq:marginal_posterior} is intractable, in particular because the dimension of $\VTheta$ grows linearly with the number of training samples. Even if the integral was analytically tractable, computing marginal moments from  $p\left(\mathcal{W}|\mathcal{D}_N\right)$ would remain challenging, irrespective of whether the weights are considered as continuous or discrete random variables. Thus approximate methods are needed. In the remainder of this paper, the prior model $p_0(\mathcal{W})$ is assumed to be fully separable and known (user-defined).

\section{Proposed EP-based training method}
\label{sec: EP_BNN}
In this section, we derive an EP-based method to approximate
\begin{eqnarray}
\label{eq:joint_posterior}
    p\left(\mathcal{W},\VTheta|\mathcal{D}_N\right) \propto p_0\left(\mathcal{W}\right) f(\VTheta,\*V^{(L)}|\mathcal{W},\*V^{(0)}) , 
\end{eqnarray}
by a surrogate distribution $q(\mathcal{W},\VTheta)$, whose marginal moments are easier to manipulate than those of $p\left(\mathcal{W},\VTheta|\mathcal{D}_N\right)$. In a traditional EP fashion, the factorization of $q(\mathcal{W},\VTheta)=q(\mathcal{W})q(\VTheta)$ will mimic the factorization of $p\left(\mathcal{W},\VTheta|\mathcal{D}_N\right)$. 

\subsection{Factor graph and factorization of the approximating distribution}

\begin{figure*}[!ht]
\centering
{\includegraphics[width=1\textwidth]{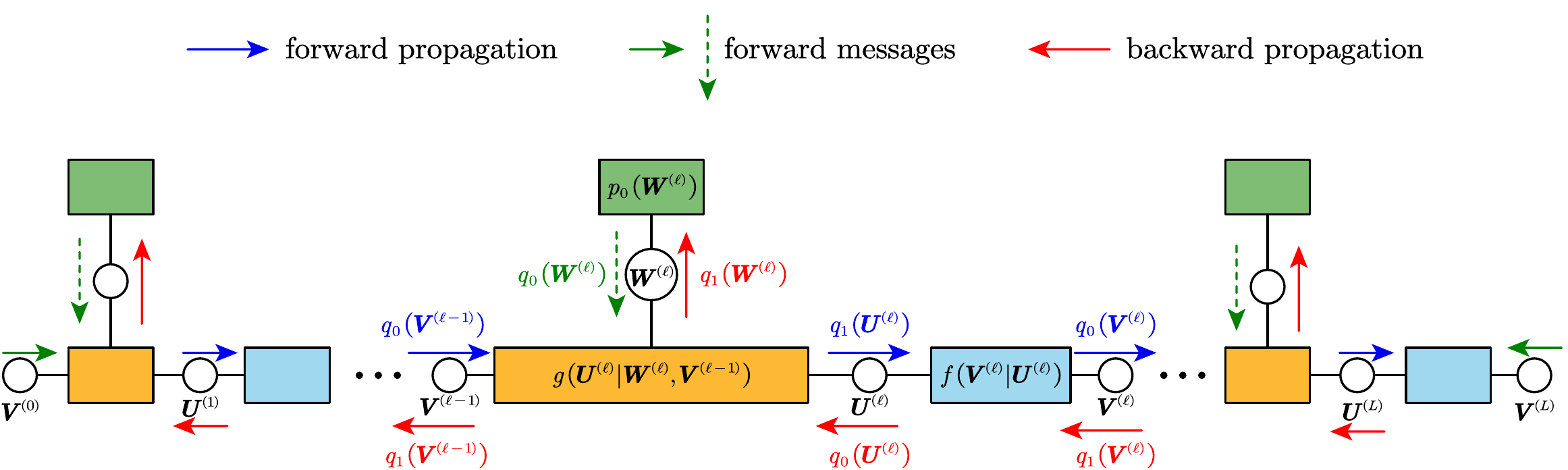}}
\vspace{-0.5cm}
\caption{Factor graph used to compute approximating distributions in layered SNNs (single timestamp). The orange (resp. blue) rectangles correspond to factor nodes of mixing (resp. activation) blocs. The dashed green arrows depict the updates needed for non-conjugate weight priors.}
\label{fig: factor_graph}
\end{figure*}

The factor graph associated with Eq. \eqref{eq:joint_posterior} is illustrated in Fig. \ref{fig: factor_graph}. Each layer of the network can be decomposed into two sub-blocks: a \textit{mixing} block followed by an \textit{activation} block. Moreover, each group of variables in $\{\*U^{(\ell)},\*V^{(\ell)},\*W^{(\ell)}\}_{\ell=1}^L$ only appears in two factor nodes. Thus, the separable approximation $q(\*V^{(L)},\*U,\*V,\mathcal{W})$
of $f(\*V^{(L)},\*U,\*V,|\mathcal{W},\*V^{(0)})p_{0}(\mathcal{W})$ is constructed as the product of two-factor distributions as follows
\begin{eqnarray}
\left\{
    \begin{array}{ll}
            q(\*U^{(\ell)}) & \propto q_0(\*U^{(\ell)})q_1(\*U^{(\ell)}), ~~ \forall \ell \\
            q(\*V^{(\ell)}) & \propto q_0(\*V^{(\ell)})q_1(\*V^{(\ell)}), ~~ \forall \ell \\
            q(\*W^{(\ell)}) & \propto q_0(\*W^{(\ell)})q_1(\*W^{(\ell)}), ~~ \forall \ell \\
    \end{array}
\right.
\end{eqnarray}
The variables in $\*V$ are the outputs of spiking activation functions and are thus binary. Thus, $q_0(\*V^{(\ell)})$ and $q_1(\*V^{(\ell)})$ consists of product of independent Bernoulli distributions. The variables in $\*U$, although potentially discrete with finite support, are considered as continuous variables and products of independent Gaussian distributions are used for $q_0(\*U^{(\ell)})$ and $q_1(\*U^{(\ell)})$. 
When the weights are assumed to be continuous (resp. discrete) variables, they are assigned a separable Gaussian (resp. categorical) prior and so are $q_0(\*W^{(\ell)})$ and $q_1(\*W^{(\ell)})$. In particular, if the weights are binary, the approximating distributions are Bernoulli distributions. 

\begin{algorithm}[tbh]
    \SetKwFunction{isOddNumber}{isOddNumber}
    \SetKwInOut{KwIn}{Input}
    \SetKwInOut{KwOut}{Output}
    \SetKwInput{Initialization}{Initialize}
     \SetKwBlock{Repeat}{repeat}{until}

    \KwIn{training dataset $\mathcal{D}_N$, weight prior $p_0(\mathcal{W})$}
    \KwOut{approximate posterior distribution $q(\mathcal{W})$}
    \algrule
    \Initialization{Initialise $q_1(\mathcal{W})$, $q_0(\VTheta)$ and $q_1(\VTheta)$.}

    \Repeat 
    {\underline{Forward pass:}\\

    \For{layer $\ell$:=1 to L}{
    Mixing block: Update $q_1(\*U^{(\ell)})$\\

    Activation block: Update $q_0(\*V^{(\ell)})$\\
    }
    \underline{Backward pass:}\\

    \For{layer $\ell$:=L to 1}{
        Activation block: Update $q_0(\*U^{(\ell)})$\\
    Mixing block: Update $q_1(\*W^{(\ell)})$ and $q_1(\*V^{(\ell-1)})$\\
    Optional: Update $q_0(\*W^{(\ell)})$

    }
    Update the global approximate posterior $q(\mathcal{W})$.
    }

\caption{EP SNN training (single batch)}
\label{Algo: EP}    
\end{algorithm}

Given the graph structure illustrated in Fig. \ref{fig: factor_graph}, the sequential updates of the approximating distributions $\{q_i(\cdot)\}_{i=0}^1$ can be achieved using a series of forward and backward passes through the factor graph, as summarized in Algo. \ref{Algo: EP}. More precisely, during the forward pass, from layer $\ell=1$ to layer $\ell=L$, the message $q_0(\*V^{(\ell-1)})$ is sent to the mixing block to update $q_1(\*U^{(\ell)})$. That message is then sent to the activation block of that layer to update $q_0(\*V^{(\ell)})$. During the backward pass, from layer $\ell=L$ to layer $\ell=1$, $q_0(\*U^{(\ell)})$ is updated and sent to the mixing block of layer $\ell$. This message is used to update $q_1(\*V^{(\ell-1)})$ and $q_1(\*W^{(\ell)})$. 

If $p_0\left(\mathcal{W}\right)$ is in the same family of parametric distributions as $q_0(\mathcal{W})$, 
then $q_0(\mathcal{W})$ can be initialised as $q_0(\mathcal{W})=p_0\left(\mathcal{W}\right)$ and does not need to be updated during the forward or backward passes. If $p_0\left(\mathcal{W}\right)$ is not Gaussian or categorical, then $q_0(\mathcal{W})$ can updated either during the forward or backward pass. Here we update it in the backward pass if needed. Typical examples where $p_0\left(\mathcal{W}\right)$ is not Gaussian arise when promoting sparsity, using for instance Bernoulli-Gaussian, Gaussian mixture or Laplace priors \cite{hernandez2015expectation,Yao2022}.

It is interesting to note that the backward pass, also used in \cite{soudry2014expectation} for single samples, is a message passing equivalent of gradient backpropagation. The convergence speed of this process can be adjusted by controlling the amount of damping during EP-based updates (see discussion in Section \ref{subsec: convergence}). In the next paragraphs, we discuss how messages are updated within mixing and activation blocks of each layer. 

\subsection{Updates for mixing blocks}
For the mixing block of layer $\ell$, we first introduce the set of variables $\*{\Phi}^{(\ell)}=\{\*U^{(\ell)},\*V^{(\ell-1)},\*W^{(\ell)}\}$. the traditional block-EP update consists of minimizing the following KL divergence
\begin{eqnarray}
KL\left[g(\*{\Phi}^{(\ell)})q_0(\*{\Phi}^{(\ell)})||q\left(\*{\Phi}^{(\ell)}\right)\right].
\end{eqnarray}
where
\begin{eqnarray}
  g(\*{\Phi}^{(\ell)}) & = & \delta\left(\*U^{(\ell)}-\*W^{(\ell)}\*V^{(\ell-1)}\right)\\
   & = & \prod_{n,i}\delta\left(u_{n,i}-\*w_{i}^{(\ell)}\*v_{n}^{(\ell-1)}\right),\nonumber
\end{eqnarray}
where, in this paragraph $\*w_{i}^{(\ell)}$ denotes the row $i$ of $\*{W}^{(\ell)}$, and for sample $n$, $\*s_{n}^{(\ell)}$ is the forward trace of layer $\ell$ and $\*v_{n}^{(\ell-1)}$ the spiking ouput of layer $(\ell-1)$.  

The product of delta functions in $g(\*{\Phi}^{(\ell)})$ does not allow easy computation of the marginal distributions associated with the tilted distribution $g(\*{\Phi}^{(\ell)})q_0(\*{\Phi}^{(\ell)})$. However, handling individual delta functions is simpler, e.g. using stochastic EP updates.
It is worth noting that each $\*w_{i}^{(\ell)}$ is involved in $N$ delta functions (because $N$ samples are considered in the factor graph) and each $\*v_{n}^{(\ell-1)}$ is involved in $V_{\ell}$ delta functions, whereas each $u_{n,i}$ is involved in a single delta function. In a similar fashion to the vanilla stochastic EP, we thus define local cavity distributions
\begin{eqnarray}
\label{eq:AEP_cavities}
    \left\{
    \begin{array}{ll}
            q_c(\*v_{n}^{(\ell)}) & \propto  q_0(\*v_{n}^{(\ell)})\left[q_{1,old}(\*v_{n}^{(\ell)})\right]^{\frac{V_{\ell}-1}{V_{\ell}}}, ~~ \forall n \\
            q_c(\*w_{i}^{(\ell)}) & \propto  q_0(\*w_{i}^{(\ell)})\left[q_{1,old}(\*w_{i}^{(\ell)})\right]^{\frac{N-1}{N}}, ~~ \forall i \\
            q_c(u_{n,i}^{(\ell)}) & = q_{0}(u_{n,i}^{(\ell)}), ~~ \forall (n,i)
    \end{array}
\right.
\end{eqnarray}
where $q_{1,old}(\cdot)$ denotes the previous estimate of $q_{1}(\cdot)$. Defining $\*{\phi}_{i,n}^{(\ell)}=\{u_{n,i},\*w_{i}^{(\ell)},\*v_{n}^{(\ell-1)}\}$, the divergence to be minimized for each delta function becomes 
\begin{eqnarray}
KL\left[\tilde{g}(\*{\phi}_{i,n}^{(\ell)})q_c(\*{\phi}_{i,n}^{(\ell)})||q\left(\*{\phi}_{i,n}^{(\ell)}\right)\right],
\end{eqnarray}
with $\tilde{g}(\*{\phi}_{i,n}^{(\ell)})=\delta\left(u_{n,i}-\*w_{i}^{(\ell)}\*v_{n}^{(\ell-1)}\right)$.
The marginal tilted moments of $u_{n,i}^{(\ell)}$ can be approximated as in \cite{soudry2014expectation}, by approximating $\eta_u= \*w_{i}^{(\ell)}\*v_{n}^{(\ell-1)}$ as a Gaussian random variable with
\begin{eqnarray}
\label{eq:q_c_eta_u}    q_c(\eta_u)=\mathcal{N}\left(\eta;\mathbb{E}_{q_c(\*{\phi}_{i,n}^{(\ell)})}\left[\eta_u\right],\textrm{Var}_{q_c(\*{\phi}_{i,n}^{(\ell)})}\left[\eta_u\right]\right).
\end{eqnarray}
The marginal moments of $\int\delta(u_{n,i}-\eta_u)q_c(\eta_u)q_c(u_{n,i})\textrm{d}\eta_u$ can then be obtained in closed-form \cite{kim2018expectation} and $q_{1}(u_{n,i})$ can finally be obtained by moment matching. 

The update for the other variables in $\*{\phi}_{i,n}^{(\ell)}$ requires an additional approximation. This is explained here for one weight of $\*w_{i}^{(\ell)}$, denoted $w_{i,j}^{(\ell)}$ but the same process applies to the other variables. Let $\*w_{i,\backslash j}^{(\ell)}$ contain all the elements of $\*w_{i}^{(\ell)}$ but $w_{i,j}^{(\ell)}$. If $v_{n,j}^{(\ell-1)}\neq 0$, $\tilde{g}(\*{\phi}_{i,n}^{(\ell)})$ can also be expressed $\tilde{g}(\*{\phi}_{i,n}^{(\ell)})=\delta(w_{i,j}^{(\ell)}-\eta_w)$, where 
\begin{eqnarray}
\label{eq:eta_w}
    \eta_w=\dfrac{u_{i,j}^{(\ell)}-\*w_{i,\backslash j}^{(\ell)}\*v_{n,\backslash j}^{(\ell-1)}}{v_{n,j}^{(\ell-1)}}.
\end{eqnarray}
In contrast to $\eta_u$, the moments of $\eta_w$ under $q_c$ are generally not tractable because the denominator in \eqref{eq:eta_w} is a random variable (except for $\ell=1$). They could be estimated, e.g. via Monte Carlo simulation, but this would significantly increase the overall computational cost of the algorithm. Instead, we fix $v_{n,j}^{(\ell-1)}$ to its mean under $q_c$ and define 
\begin{eqnarray}
\label{eq:eta_w_star}
    \eta_w^*=\dfrac{u_{i,j}^{(\ell)}-\*w_{i,\backslash j}^{(\ell)}\*v_{n,\backslash j}^{(\ell-1)}}{\mathbb{E}_{q_c(\*{\phi}_{i,n}^{(\ell)})}\left[v_{n,j}^{(\ell-1)}\right]}.
\end{eqnarray}
The distribution of $\eta_w^*$ is then approximated using
\begin{eqnarray}
\label{eq:q_c_eta_w}
    q_c(\eta_w^*)=\mathcal{N}\left(\eta;\mathbb{E}_{q_c(\*{\phi}_{i,n}^{(\ell)})}\left[\eta_w^*\right],\textrm{Var}_{q_c(\*{\phi}_{i,n}^{(\ell)})}\left[\eta_w^*\right]\right). 
\end{eqnarray}
In a similar fashion to $u_{i,n}^{(\ell)}$, the marginal moments of $\int\delta(w_{i,j}^{(\ell)}-\eta_u)q_c(\eta_u)q_c(w_{i,j}^{(\ell)})\textrm{d}\eta_u$ can be obtained in closed-form and the local approximation $q_{1,n}(w_{i,j}^{(\ell)})$ can then be obtained by moment matching. 

The local marginal tilted distributions can be computed independently for each delta function factor. It is possible to update $q_1$ and recompute the cavities $q_c$ after approximation of each of the $N V_{\ell}$ factors (vanilla SEP), or groups of factors (parallel SEP with mini-batches) but to reduce the number of sequential updates, we used average EP, where $q_1$ is updated after having computed the local approximations of the $N V_{\ell}$ factors.
As mentioned previously, $\*w_{i}^{(\ell)}$ is involved in $N$ delta functions, each leading to, after moment matching, a new approximation $q_{1,n}(\*w_{i}^{(\ell)})$. These approximations are combined to lead to the new approximation 
\begin{eqnarray}
\label{eq:q1_w}
    q_{1,new}(\*w_{i}^{(\ell)}) \propto \prod_{n=1}^N q_{1,n}(\*w_{i}^{(\ell)}).
\end{eqnarray}
Similarly, each  $\*v_{n}^{(\ell-1)}$ is involved in $V_{\ell}$ delta functions, each leading, after moment matching, to new approximations $q_{1,j}(\*s_{n}^{(\ell)})$ and $q_{1,j}(\*v_{n}^{(\ell)})$. These approximations are combined to lead to the new approximation 
\begin{eqnarray}
\label{eq:q1_v}
    q_{1,new}(\*v_{n}^{(\ell)}) & \propto \prod_{j=1}^{V_{\ell}} q_{1,j}(\*v_{n}^{(\ell)}).
\end{eqnarray}

\begin{algorithm}[tbh]
    \SetKwFunction{isOddNumber}{isOddNumber}
    \SetKwInOut{KwIn}{Input}
    \SetKwInOut{KwOut}{Output}
    \SetKwInput{Initialization}{Initialize}
     \SetKwBlock{Repeat}{repeat}{until}

    \KwIn{Approximating factors $q_0(\*{\Phi}^{(\ell)}),q_1(\*{\Phi}^{(\ell)})$}
    \KwOut{Updated approximate factor $q_{1,new}(\*{\Phi}^{(\ell)})$}
    \algrule
    \Initialization{Set $q_{1,old}(\*{\Phi}^{(\ell)})=q_1(\*{\Phi}^{(\ell)})$.}

    \Repeat 
    {
    Compute the cavity distributions $q_c$ in \eqref{eq:AEP_cavities}.\\

    \For{sample n:=1 to N}{
        \For{output size i:=1 to $V_{\ell}$}{
        Update $q_{1,new}(u_{n,i})$\\
            \For{input size j:=1 to $V_{\ell-1}$}{
            Compute $q_{1,n}(w_{i,j}^{(\ell)})$ and $q_{1,j}(v_{n,i}^{(\ell)})$\\
    }
    }
    }
    Update $q_{1,new}$ using Eqs. \eqref{eq:q1_w} and \eqref{eq:q1_v} and set $q_{1,old}=q_{1,new}$. 
    }

\caption{Average EP for update of the mixing block}
\label{Algo: AEP}    
\end{algorithm}

SEP/AEP is an iterative method that requires successive updates of $q_{1,new}$ to converge. The pseudo-code is provided in Algo. \ref{Algo: AEP}. It should be noted that the original AEP method updates all the approximating factors or $q_1(\*{\Phi}^{(\ell)})$. However, $q_{1,new}(u_{n,i})$ is not used in the cavities $q_c(\cdot)$ in Eq. \eqref{eq:AEP_cavities}. Thus, in the backward pass, line 5 of Algo. \ref{Algo: AEP} can be omitted, since $q_{1,new}(u_{n,i})$ will be updated during the next backward pass. Similarly, in the forward pass, it is possible to fix $q_1(\*W^{(\ell)},\*V^{(\ell)})$ and only update $q_{1,new}(u_{n,i})$, which requires a single iteration of the outer loop of Algo. \ref{Algo: AEP}. 

\subsection{Updates for activation blocks}
At the layer $\ell$, the update consists of minimizing the following KL divergence
\begin{eqnarray}
KL\left[f\left(\*V^{(\ell)}|\*U^{(\ell)}\right)q_0(\*V^{(\ell)},\*U^{(\ell)})||q\left(\*V^{(\ell)},\*U^{(\ell)}\right)\right].
\end{eqnarray}
For the activation functions considered in Section \ref{subsec: Spiking neurons under Spike Response Mode}, the first argument of the KL divergence (called the tilted distribution), as well as $q(\cdot)$, are fully separable. Thus, this problem reduces to solving $N V_{\ell}$ independent KL minimization problems with 1D distributions, which reduces to computing the mean and variances of the marginal tilted distributions. For the sigmoid activation function, the moments of the marginal distribution for the membrane potentials have been discussed in\cite{daunizeau2017semi}. For the Heaviside activation, the moments can be computed using \cite[Chap. 3.1]{Minka_thesis2001}. Using the moments of the tilted distributions, one can then compute the marginal moments of $q_1\left(\*V^{(\ell)}\right)$ and/or $q_1\left(\*U^{(\ell)}\right)$, potentially under additional constraints and damping (see Section \ref{subsec: convergence}). 

\section{Algorithmic considerations}
\label{sec: Algorithmic considerations}

\subsection{Inference with large datasets}
\label{subsec: Inference with large datasets}
As discussed in Section \ref{sec: EP_BNN}, if all the training data can be processed in a single batch, the proposed method consists of a main loop running EP-based message passing forward and backward through the network layers. To handle mini-batches, here we propose to use SEP \cite{li2015stochastic} and partition the training dataset $\mathcal{D}_N$ into a set of $B$ mini-batches $\mathcal{D}_N = 
\{\mathcal{D}_b\}_{b=1}^B$, with $M=N/B$ training samples per batch. 

Let $q(\mathcal{W})$ be the final approximation of the marginal posterior of the network weights. Using the SEP formalism, $q(\mathcal{W})$ can be factorized as
\begin{equation}
    q\left(\mathcal{W}\right) \propto \left[q_1\left(\mathcal{W}\right)\right]^B q_0\left(\mathcal{W}\right),
\label{Eq: SEP}    
\end{equation}
where $q_1\left(\mathcal{W}\right)$ represents the average contribution of each batch to the posterior distribution of the weights. The SEP method we use here starts from an initial guess for $q_1\left(\mathcal{W}\right)$, which will be updated sequentially using the $B$ batches. It is often required to process the $B$ batches several times (i.e. runs multiple epochs of SEP), however this does not lead to over-fitting as SEP automatically accounts for the total amount of data while updating $q_1\left(\mathcal{W}\right)$. 
To process a batch $b$, SEP first creates a cavity distribution
\begin{eqnarray}
    q_c\left(\mathcal{W}\right) & \propto &\left[q_{1,old}\left(\mathcal{W}\right)\right]^{B-1} q_0\left(\mathcal{W}\right)\nonumber\\
    & \propto & q\left(\mathcal{W}\right)/q_{1,old}\left(\mathcal{W}\right).
\label{Eq: SEP_cavity}    
\end{eqnarray}
This cavity distribution is then used as a local prior model for $\mathcal{W}$ to be trained with $\mathcal{D}_b$, essentially replacing $p0\left(\mathcal{W}\right)$ by $q_c\left(\mathcal{W}\right)$ in Section \ref{sec: EP_BNN}. This leads to a local approximated posterior distribution denoted $q_b(\mathcal{W})$, which will be in the same parametric family of distributions as $q(\mathcal{W})$. In that case, $q_{1,b}(\mathcal{W})\propto q_{b}(\mathcal{W})/q_c\left(\mathcal{W}\right)$ can be used to update $q_1\left(\mathcal{W}\right)$ via 
\begin{eqnarray}
\label{eq:update_SEP}
 q_{1,new}\left(\mathcal{W}\right) = \left[q_{1,old}\left(\mathcal{W}\right)  \right]^{1-1/B} \left[q_{1,b}\left(\mathcal{W}\right)\right]^{1/B}.
\end{eqnarray}
The update of $q_0(\mathcal{W})$ can then be performed after each batch, in a similar fashion to Section \ref{sec: EP_BNN}, but using $\left[q_{1,old}\left(\mathcal{W}\right)\right]^{B-1}$ as cavity distribution. 
The pseudo-code of the proposed SEP method is summarized in Algo. \ref{Algo: SEP}. 
\begin{algorithm}[tbh]
    \SetKwFunction{isOddNumber}{isOddNumber}
    \SetKwInOut{KwIn}{Input}
    \SetKwInOut{KwOut}{Output}
    \SetKwInput{Initialization}{Initialize}
     \SetKwBlock{Repeat}{repeat}{until}

    \KwIn{training dataset $\mathcal{D}_N$, weight prior $p_0(\mathcal{W})$, batch size $M$}
    \KwOut{approximate posterior distribution $q(\mathcal{W})$}
    \algrule
    \Initialization{Set $B=N/M$ and initialise $q_1(\mathcal{W})$.}

    \Repeat 
    {Randomize input data into $B$ batches ($\mathcal{D}_N = 
\{\mathcal{D}_b\}_{b=1}^B$).\\

    \For{batch b:=1 to B}{
    Compute the cavity distribution $q_c(\mathcal{W})$ in \eqref{Eq: SEP_cavity}\\

    Run Algo. \ref{Algo: EP} using the data $\mathcal{D}_b$ and $q_c(\mathcal{W})$ as weight prior. The ouput of the algorithm is denoted $q_{b}(\mathcal{W})$.\\
    
    Update $q_1(\mathcal{W})$ using \eqref{eq:update_SEP} \\
    Optional: Update $q_0(\mathcal{W})$
    }
    Update the global approximate posterior $q(\mathcal{W})$.
    }

\caption{Stochastic EP for large datasets}
\label{Algo: SEP}    
\end{algorithm}

\subsection{Convergence}
\label{subsec: convergence}
While convergence of the proposed method cannot be ensured, a series of tools are available to stabilize Bayesian learning. Optimal settings will be architecture-dependent but we report here a few empirical results to help future practitioners. Careful initialization of approximating factors can help stabilize the algorithm especially during the first epochs or first batches. The shallow architectures considered in this paper were trained from scratch, but for deeper networks, pre-training might help converge to better solutions. 


Another important tool typically used in EP-based method is damping, which consists of limiting the changes of approximating distributions between successive updates (using weighted geometric averages) \cite{hernandez2015expectation}. In this work, damping is used in the outer loop of  SEP in Algo. \ref{Algo: SEP} and also  used in the inner loop of AEP in Algo. \ref{Algo: AEP}. The damping parameter used in SEP is set to be $1/B$  following \cite{li2015stochastic} (see \eqref{eq:update_SEP}), and set to 0.7 in AEP (Algo. \ref{Algo: AEP}). 

We also observed that regions of the networks which are rarely activated can be particularly prone to variances becoming negative if not constrained. This often occurs within EP and can be mitigated by adding constraints to the KL minimization problems \cite{hernandez2015expectation}. We observed that, especially during the early epochs, bounding the variances of Gaussian approximating factors might not be sufficient. Thus we propose to bound the two natural parameters of the Gaussian factors. The derivation of the iterative algorithm is similar to that in \cite{hernandez2015expectation} and omitted here due to space constraints, together with the constrained minimization for categorical distributions. 

\begin{table*}[!ht]
\centering
\resizebox{0.8\textwidth}{!}{
\begin{tabular}{lcccccc}
\multicolumn{2}{c}{} & \multicolumn{1}{c}{MLE} &  \multicolumn{4}{c}{EPSNN}  \\
\cmidrule(r){3-3}\cmidrule(l){4-7}
 \multicolumn{2}{c}{Weight type} & Continuous    & Binary  & Binary    &   Continuous & Continuous  \\
 \multicolumn{2}{c}{Layer activation} &  Heaviside&Heaviside & Sigmoid & Heaviside & Sigmoid\\
\cmidrule{1-7}
\multirow{2}{*}{Accuracy} &
 $N=10^3$& {0.97/0.84}&  {0.96/0.76}  &{0.97/0.80} &{0.98/0.84}& {1.00/0.84} \\
& $N=10^4$ & {0.92/0.88}  & {0.89/0.86} &{0.90/0.86} & {0.93/0.86} & {0.93/0.87}\\
\cmidrule{1-7}
\multirow{2}{*}{Losses} &
 $N=10^3$& {0.04/0.11}&  {0.06/0.19}  &{0.06/0.13} &{0.07/0.12}& {0.07/0.12} \\
& $N=10^4$ &  {0.07/0.09}  & {0.09/0.12} & {0.08/0.11}&{0.06/0.12}& {0.06/0.09}\\
\cmidrule{1-7}
\end{tabular}}
\caption{Comparison of classification accuracies and losses, using a 1-layer architecture ($784$ inputs, $10$ outputs). The first (resp. second) values are the metrics computed on the test (resp. test) data. EPSNN is run with batches of $M=10^3$ samples.}
\label{Table: comparison_determinstic_probabilistic}
\end{table*}

\subsection{Message passing at inference time}
\label{subsec: testing}
While we introduced our EP method for training SNNs, The same EP framework can be used at inference time, to predict the marginal means and variances of the network outputs without resorting to Monte Carlo sampling, e.g. without generating multiple random realizations of the weights nor passing data multiple times through the network as in \cite{jang2021bisnn} to obtain ensemble estimates. After training, it is possible to consider a factor graph similar to that in Fig. \ref{fig: factor_graph}, where all the approximating factors but those associated with the network weights are set to uninformative messages (uniform distributions for the Bernoulli variables and Gaussians with large variances for the Gaussian factors). The initial prior factors for the weights can then be replaced by the marginal posteriors obtained after training. At inference time, it is sufficient to run a single forward pass.

\section{Experiments}
\label{sec: Experiments}
This section evaluates the performance of SNNs trained by our method, referred to as \textit{EPSNN}, and a benchmark method for a classification and a regression task. For quantitative comparisons, we consider as baseline the PyTorch implementation of the method in \cite{eshraghian2021training} (referred to as maximum likelihood estimation (MLE)), which estimates continuous weights and relies on Heaviside activation functions. The regression experiment is similar to that considered in \cite{jang2021bisnn} and allow us visual comparisons with the BiSNN method.

\subsection{Performance metrics and losses}
\label{subsec: metrics}
For consistency purposes, MLE and EPSNN are trained with similar output likelihoods. More precisely, for the classification problem, the binary cross-entropy (BCE) loss is used for MLE, i.e., 
\begin{eqnarray}
\textrm{BCE}_{i,n}=y_{i,n}\log(\sigma(u_{i,n}))+(1-y_{i,n})\log(1-\sigma(u_{i,n})),\nonumber
\end{eqnarray}
where $u_{i,n}$ is the membrane potential of the $i$th
output of the network, for the $n$th sample. 

Since EPSNN does not estimate a single set of weights but their marginal distribution, the loss reported is the posterior expected BCE (PeBCE)
\begin{eqnarray}
\textrm{PeBCE}_{i,n}=\mathbb{E}_{q(u)}\left[  \textrm{BCE}_{i,n}\right].\nonumber
\end{eqnarray}
where the expectation is taken with respect to the variational approximation of the distribution of $u_{i,n}$ (see Section \ref{subsec: testing}). 
We also report the classification accuracy, where the output with the lowest BCE/PeBCE is selected as estimated class. 

For the regression problem, we use the mean squared error (MSE)
between the membrane potential of the output layer and the ground truth as loss for the MLE-based method. While we presented spiking likelihoods in \eqref{Eq: spiking emission function}, replacing them by a Gaussian likelihood for the output layer is straightforward and this is the option adopted here for regression problems. For EPSNN, we report the MSE 
 computed using the mean of $q(u_{i,n})$ at inference time.

\subsection{Image classification}

We first consider an image classification task using the MNIST dataset \cite{deng2012mnist}. Spiking input images are generated by mapping linearly the original $28\times28$-pixel image intensities into firing rates between $0$ and $1$, using a single timestamp (no temporal sequences). The training sets consists of up to $10^4$ frames, and all the digits are represented with equal probabilities. The test set consists of $10^4$ images. Although MLE can only estimate continuous weights using the Heaviside activation, here we compare different versions of EPSNN, with Heaviside and sigmoid activations, binary (in $\{-1,+1\}$) and continuous weights.

\begin{figure}[htb!]
\centering
\includegraphics[width=0.5\textwidth]{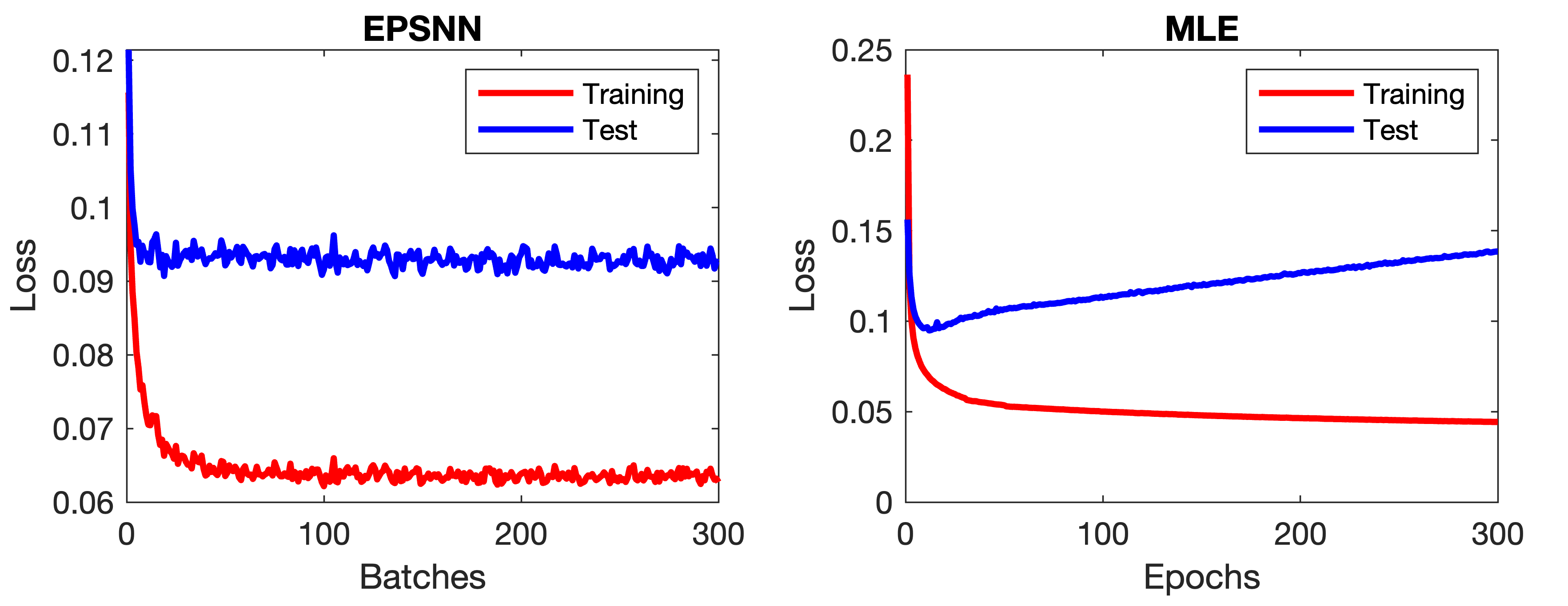}
\vspace{-0.8cm}
\caption{Convergence analysis of MLE-based optimization and EPSNN (continuous weights, sigmoid activation), for the MNIST classification problem with $N=10^4$ training data. EPSNN generally converges faster than MLE. For visualisation purposes, EPSNN is stopped after the first $300$ batches ($30$ epochs) and we visualise the first $300$ MLE epochs.}
\label{fig: result_convergence}
\end{figure}

\begin{figure}[htb!]
\centering
\includegraphics[trim={2cm 2cm 8cm 1cm },width=0.5\textwidth]{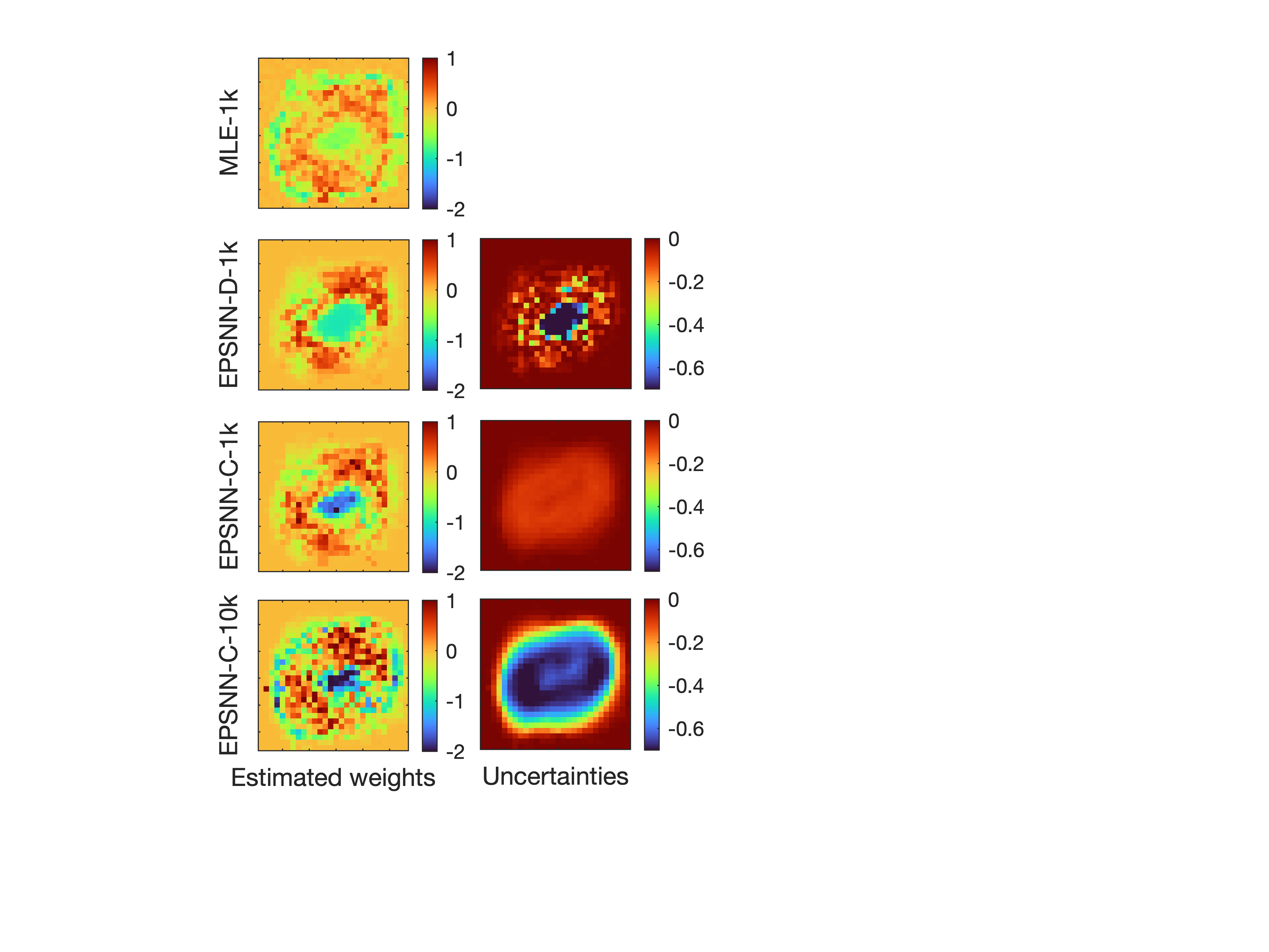}
\vspace{-0.8cm}
\caption{Estimated weights associated with the output of the '0' digit (left) and associated log-uncertainties (right), reshaped into $28\times 28$ images.}
\label{fig: result_MNIST_weights1}
\end{figure}

Table \ref{Table: comparison_determinstic_probabilistic} compares the metrics introduced in Section \ref{subsec: metrics} for a simple 1-layer architecture with one-hot encoded outputs (10 neurons corresponding to the 10 classes), obtained by MLE and EPSNN. As can be observed in Fig. \ref{fig: result_convergence}, the MLE-based method uses the Adam optimizer and is prone to overfitting. Thus, early stopping is adopted here and the optimization is stopped when the test loss start increasing after the initial decrease, for all results reported in this paper. In contrast, EPSNN does not rely on early stopping. While it also tends to overfit the training data, the test metrics do not diverge as SEP iterates. This figure also shows that the results improve with more than one pass through the data, which is not possible using the EP-based methods reported in Section \ref{sec: Introduction}. Table \ref{Table: comparison_determinstic_probabilistic} illustrates how the test metrics generally improve as more training data is available, for both MLE and EPSNN. It is interesting to observe that the performance of EPSNN with continuous weights is on a par with the MLE-based method, while also providing weight uncertainties after training (see Fig. \ref{fig: result_MNIST_weights1}). As expected, the performance generally degrades with quantized weights, though not dramatically so given their heavily quantized nature. It is worth also highlighting that the EPSNN results are generally better with the sigmoid activation than with the Heaviside activation, which we believe is mostly due to the numerical/rounding errors when computing messages in the activation blocks. This is more obvious when using quantized weights.

Fig. \ref{fig: result_MNIST_weights1} depicts the $784$ estimated weights associated with the output of the '0' digit, reshaped into $28\times 28$ images. This figure presents the weights obtained by MLE with 1000 training data (MLE-1k), EPSNN with discrete (resp. continuous) weights, sigmoid activation and 1000 training data (EPSNN-D-1k, resp. EPSNN-C-1k), and EPSNN with continuous weights, sigmoid activation and 10000 training data (EPSNN-C-10k). For EPSNN with discrete weights, the posterior mean is used as point estimate. All methods tend to agree that the weights associated with the periphery of the images should be small, i.e. have limited impact on the network decision. The right-hand side columns of Fig. \ref{fig: result_MNIST_weights1} depicts the estimated marginal posterior variances of the weights (in log-scale). The large variances in the periphery of the images coincides with the prior variances and confirm that the training data do not provide additional information to estimate those weights. As expected, the variance of the weights decreases for the informative weights as more training data is available.

\begin{table}[!ht]
\small
\centering
\begin{tabular}{cccc}
\multicolumn{1}{c}{} & MLE &  \multicolumn{2}{c}{EPSNN}  \\
\cmidrule(r){2-2}\cmidrule(l){3-4}
Prior &   $-$  & Gaussian & Bernoulli-Gaussian\\
\cmidrule{1-4}
Acc. &   1.00/0.85  & 0.90/0.83 & 0.92/0.83\\
\cmidrule{1-4}
Loss &0.02/0.09 & 0.12/0.14 & 0.11/0.14 \\
\cmidrule{1-4}
\end{tabular}
\caption{Comparison of classification accuracies and losses, using a 2-layer (784-120-10) architecture. The first (resp. second) values are the metrics computed on the test (resp. test) data.}
\label{Table: comparison_MNIST_1_2_layers}
\end{table}

Table \ref{Table: comparison_MNIST_1_2_layers} reports the performance metrics obtained by MLE and EPSNN on the same classification problem and $N=10^3$, but using a 2-layer architecture with $120$ neurons in the hidden layer. Here we report only the EPSNN results obtained with the Heaviside activation function for comparison with MLE but investigate the impact of the weight prior. More precisely, we considered a first prior model $p_{0,1}(\mathcal{W})$ where all weights share the same standard Gaussian prior. In the second model $p_{0,2}(\mathcal{W})$, all the weights are assigned a Bernoulli-Gaussian prior with the same weight for the Bernoulli component and for the standard Gaussian component. The aim is for the network to identify features in the first layer that depend on a reduced number of input pixels and such that each of the $10$ outputs of the network depends on a reduced number of features, leading to a sparse combination of sparse features. MLE, if tuned properly to minimize the test loss, performs marginally better than EPSNN. Moreover, the sparse prior model does not significantly degrade the output test metrics, and lead to better training metrics. 

\begin{figure}[htb!]
\centering
\includegraphics[width=0.47\textwidth]{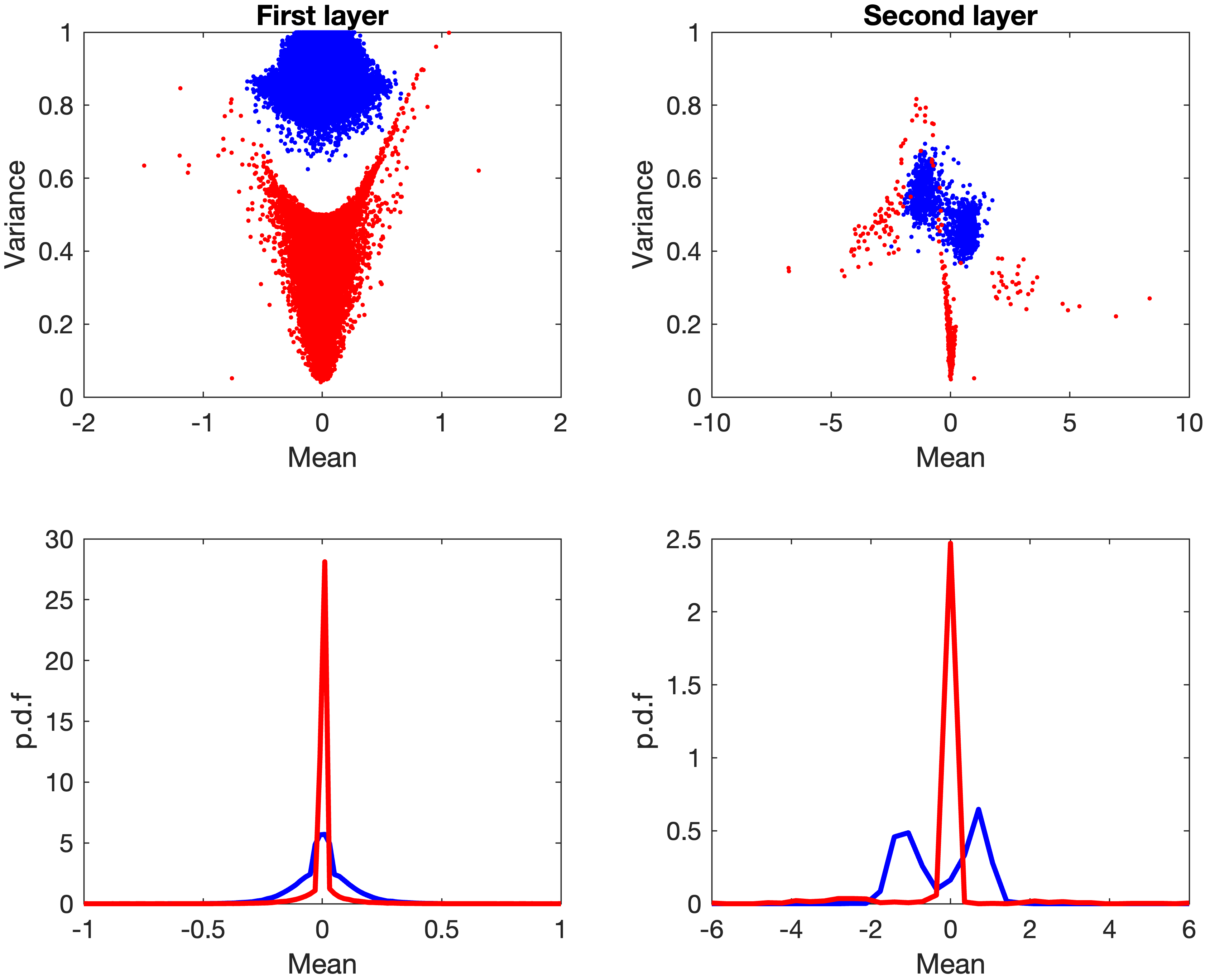}
\caption{Estimated weight distributions for the MNIST classification problem using a 2-layer architecture. The top row depicts the marginal means and variances for the first (left) and second (right) layer, for the prior model $p_{0,1}(\mathcal{W})$ (blue) and $p_{0,2}(\mathcal{W})$ (red). The bottom row depicts the distributions of the weight posterior means using the $p_{0,1}(\mathcal{W})$ (blue) and $p_{0,2}(\mathcal{W})$ (red).}
\label{fig: result_MNIST_weights2}
\end{figure}

The weight distributions estimated by EPSNN are depicted in Fig. \ref{fig: result_MNIST_weights2}. This figure confirms that the sparse weight model indeed provides on average smaller means and variances. While for continuous weights, our EP method relies on Gaussian approximating distributions, it is still possible to estimate marginal \textit{probabilities of activation} of the weights when a spike-and-slab prior is considered. More precisely, as in \cite{hernandez2015expectation,Yao2022}, at the last update of $q_0(\mathcal{W})$, one can store as approximating posterior distribution of the weights the corresponding tilted distribution, which is a Bernoulli-Gaussian distribution. The weight of the mixture can be interpreted as a probability of activation. 
\begin{figure}[htb!]
\centering
\includegraphics[trim={1.5cm 3cm 1cm 0 },width=0.5\textwidth]{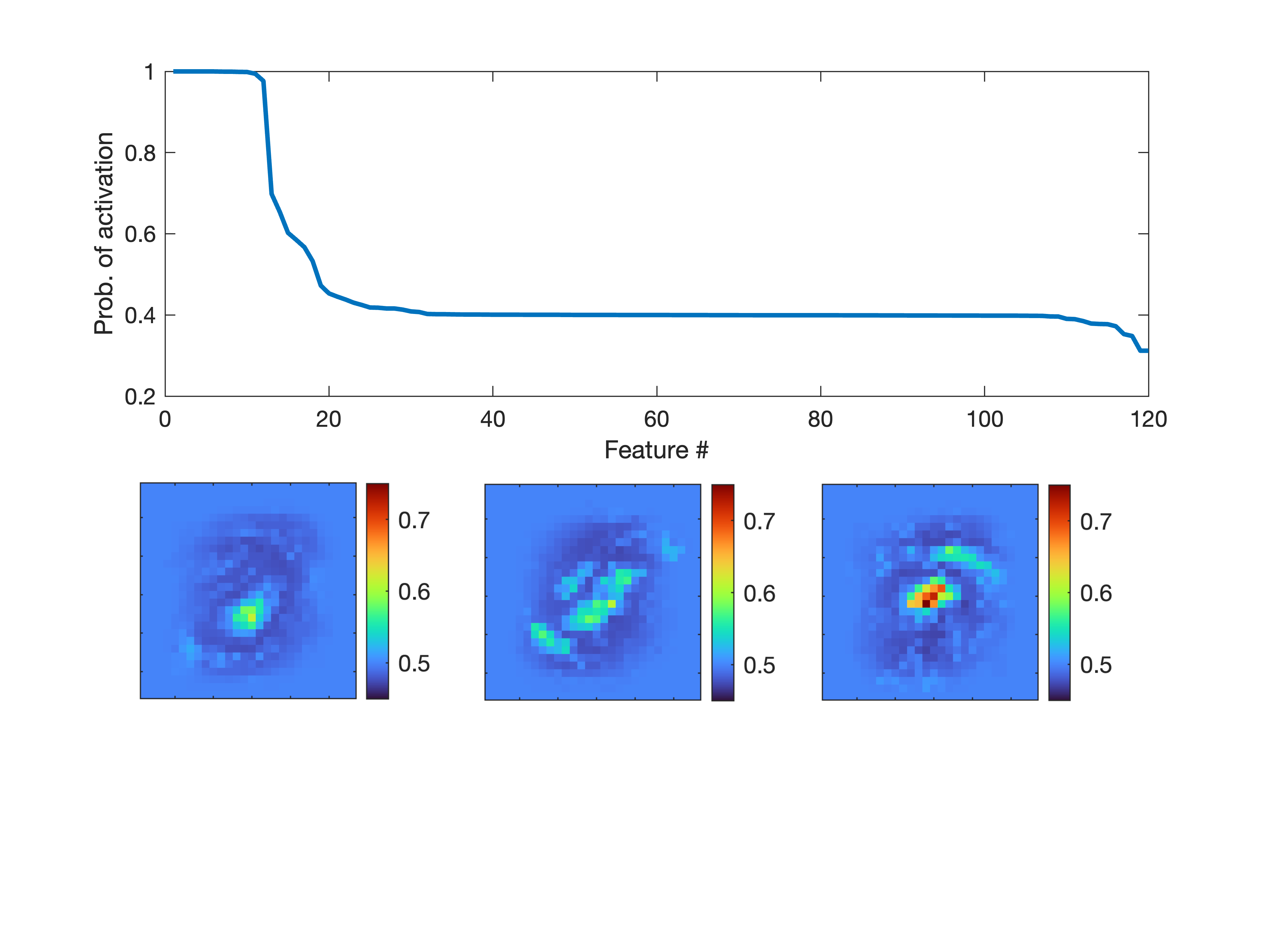}
\vspace{-0.8cm}
\caption{Top: posterior probability of activation of the 120 features of the hidden layer, for the '0' digit, sorted in decreasing order. Bottom: posterior probability of activation of the $28 \times 28$ weights of the first layer, for the three most likely features of the '0' digit.}
\label{fig: result_MNIST_weights3}
\end{figure}
The top subplot of Fig. \ref{fig: result_MNIST_weights3} depicts those probabilities for the weights of the $120$ features of the hidden layers, linked to the identification of the '0' digit (features sorted by decreasing probability of activation). It shows that about 15 features should be used with high marginal probability to identify that digit. Despite the prior model promoting $50\%$ sparsity levels, most features have a lower probability of activation, which indicates that the network tends to remove the contribution of some features. The bottom row of Fig. \ref{fig: result_MNIST_weights3} depicts probabilities of activation (at the pixel level, i.e., using the weights of the first layer) of the three features with the highest marginal probability of activation. First, it can be observed that these images share common spatial features, in line with the '0' digit and that for the pixels at the periphery of the images, the probability of activation is $50\%$, which corresponds to the prior belief. Those features seem to capture variability in the handwritten digits of the images.

\subsection{Regression}

\begin{figure}[htb!]
\centering
\includegraphics[width=0.48\textwidth]{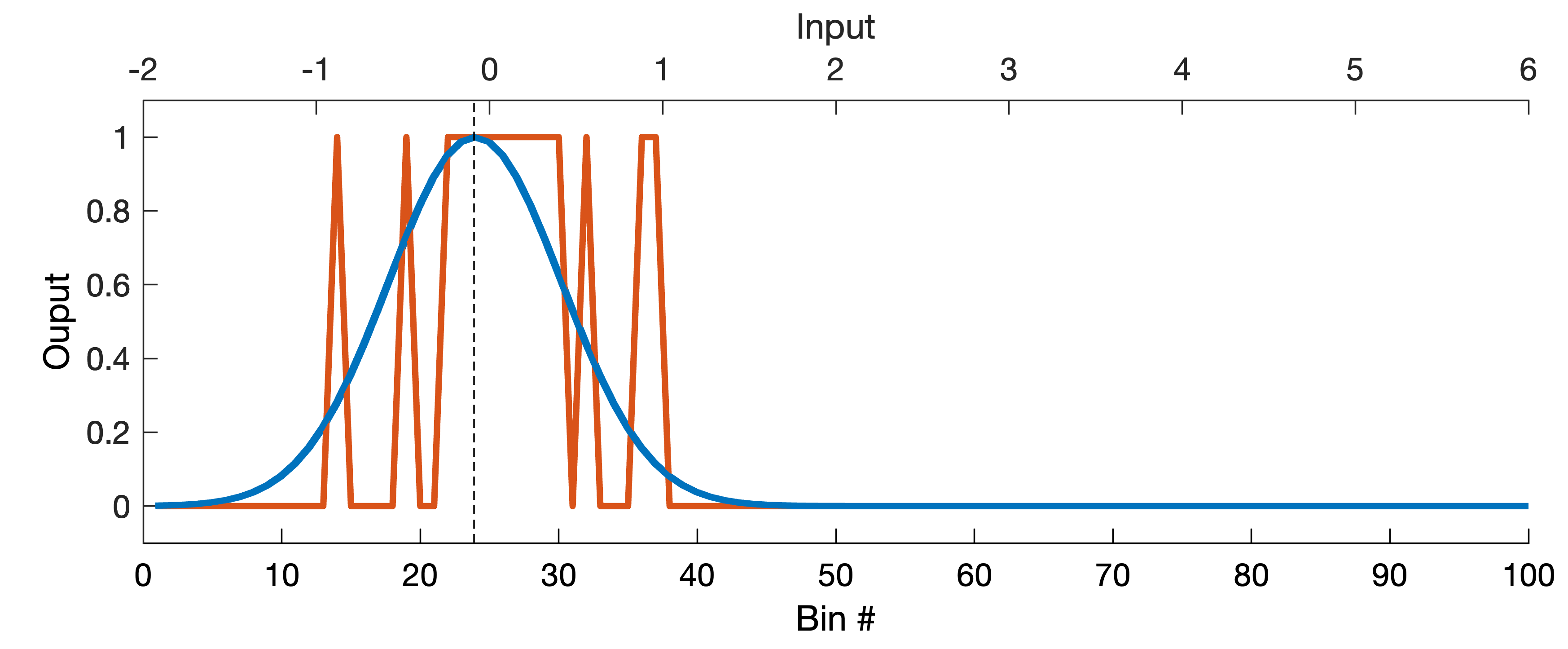}
\vspace{-0.8cm}
\caption{Example of population encoding for the regression problem. For each input $x\in [-1,5]$ ($x=-0.15$ here), a vector of $100$ spiking rates are created, with a peak centered at $x$ (blue line). Random spiking outputs are then generated from Bernoulli distributions (red line) and fed to the network.}
\label{fig: encoding_regression}
\end{figure}

\begin{figure}[htb!]
\centering
\includegraphics[width=0.5\textwidth]{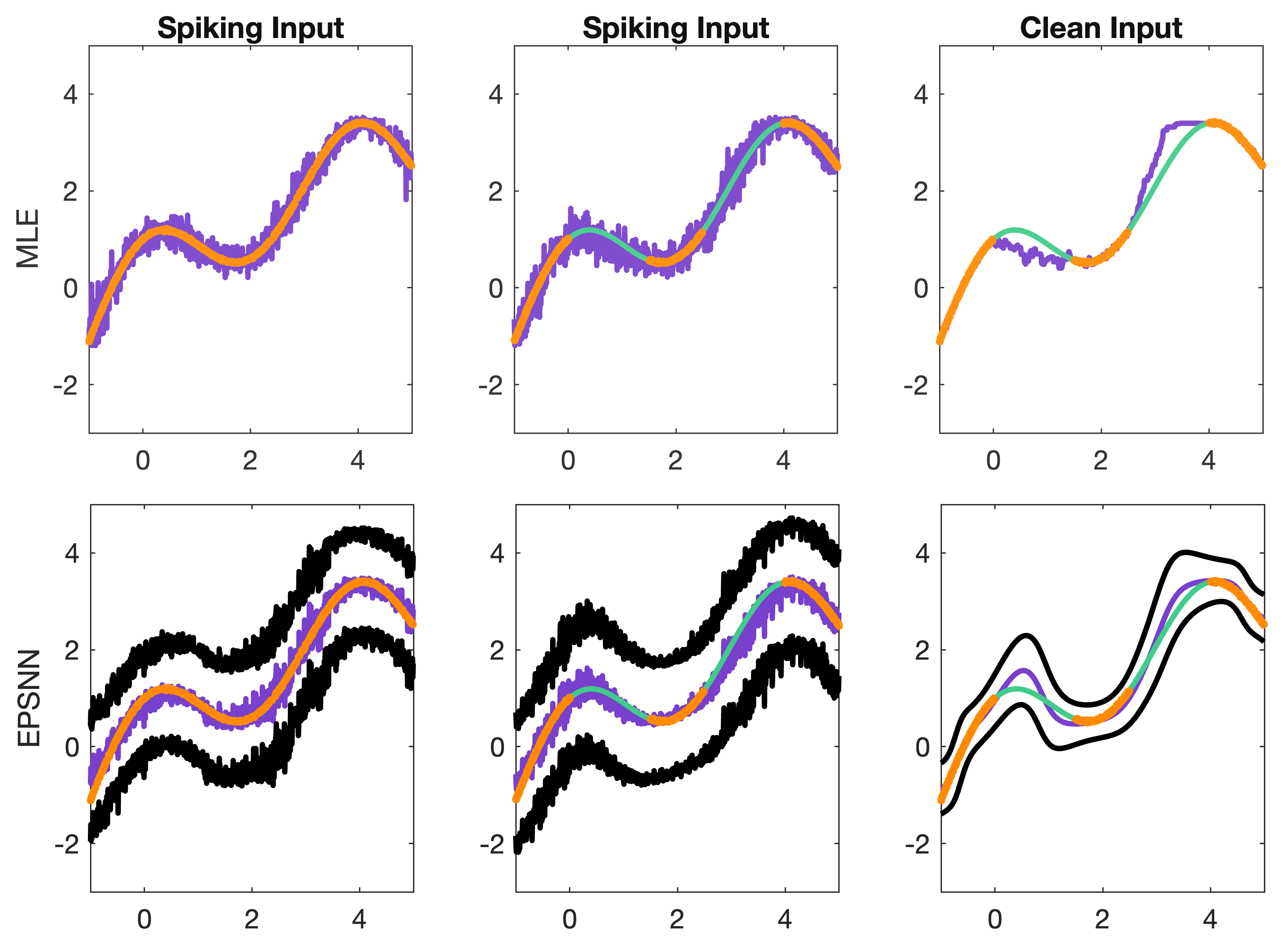}
\vspace{-0.8cm}
\caption{Example of 1D regression with population encoding using MLE (top) and EPSNN (bottom). Green lines depict the ground truth, the training data appear in orange and the estimated outputs are in violet. For EPSNN, the black lines indicate the $\pm 1$ posterior standard deviation intervals. In the left-hand side column, the training data span the whole regression domain while gaps appear otherwise.}
\label{fig: result_regression}
\end{figure}

Here we investigate a regression task similar to that in \cite{jang2021bisnn}, but using a single timestamp instead of spiking time series. We used sets of $N=600$ training samples, to solve the 1D regression problems depicted in Figure \ref{fig: result_regression}. The output $y\in \mathbb{R}$ is related to the input $x\in [-1;5]$ via the nonlinear function $f(x)=x - 0.1 x^2 + \cos(\pi x / 2)$ and is corrupted by additive zero-mean white Gaussian noise with variance $10^{-4}$. Moreover, each training/test data $x$ is encoded onto $100$ input neurons with population encoding \cite{Eliasmith2002}. As illustrated is Fig. \ref{fig: encoding_regression}, raised cosine filters are created to generate 100-dimensional vectors of spiking rates which are then used to generate random spiking vectors. This sampling process increases uncertainties in the training and test data. The first dataset $\mathcal{D}_1$ consists of input samples uniformly distributed in $[-1,5]$ (see left column of Fig. \ref{fig: result_regression}). The second dataset $\mathcal{D}_2$ consists of training samples $x$ uniformly sampled in $[-1,0]\cup[1.5,2.5]\cup[4,5]$ to investigate the ability of the network to predict missing data (see central column of Fig. \ref{fig: result_regression}). The third dataset is similar to $\mathcal{D}_2$ but the networks are fed with the noise-free spiking rates instead of the spiking signals (right column of Fig. \ref{fig: result_regression}).   

The SNN architecture consists of one hidden layer with Heaviside activation function, using $100$-$128$-$1$ neurons. For this regression task, the output layer activation function is replaced by a Gaussian model $f(v|u)$ with fixed variance matching that used to generate the training data. EPSNN uses a standard Gaussian prior for all the weights. 

The prediction results obtained by MLE and EPSNN are shown in Fig. \ref{fig: result_regression} and the training/test MSEs are compared in Table \ref{Table: comparison_regression}. These results first show that EPSNN performs marginally better for test data when they coincide with the training set ($\mathcal{D}_1$). We also observed that it is generally the case also for $\mathcal{D}_2$ and $\mathcal{D}_3$ where training data is missing, although the MLE results there depend on when the algorithm is stopped. Conversely, EPSNN provides satisfactory interpolation results and uncertainties in line with the (unobserved) ground truth in the $[0,1]\cup[2.5,4]$ domain. Interestingly, EPSNN provides lower uncertainties when the network is fed noise-free spiking rates instead of spiking signal (see Fig. \ref{fig: result_regression}), which is expected as the training data present different levels of uncertainty. 

\begin{table}[!ht]
\small
\centering
\begin{tabular}{cccc}
  &   Spiking (full)  & Spiking & Noise-free\\
\cmidrule{1-4}
MLE &   0.002/0.036  & 0.002/0.046 & 0.001/0.071\\
\cmidrule{1-4}
EPSNN &0.031/0.032 & 0.025/0.040 & 0.005/0.026\\
\cmidrule{1-4}
\end{tabular}
\caption{Fitting performance of MLE-based and EPSNN-based regression, using a 2-layer (100-128-1) architecture and $N=600$ training data. The first (resp. second) values are the MSEs computed on the test (resp. test) data.}
\label{Table: comparison_regression}
\end{table}

\section{Conclusions and Future Work}
\label{sec: Conclusions and Future Work}
We have proposed a unifying framework based on Expectation-Propagation (EP) for training spiking neural networks. In contrast to existing EP-based methods for Bayesian neural network training, we have proposed a combination of stochastic and average EP updates to handle mini-batches. Due to its message-passing interpretation, the proposed framework is modular and flexible, can be used for continuous and discrete weights, as well as deterministic and stochastic spiking networks. We conducted experiments with architectures including one and two hidden layers and the results are in line with those obtained using existing methods. We also observed that the algorithm generally converges rapidly without the need of a large number of passes through the training data. 
The neuron models and architectures investigated in this paper need to be generalized in the future, in particular for use with recurrent neurons and to include feedback mechanisms \cite{jang2019introduction,kaiser2020synaptic} and winner-take-all units \cite{oster2009computation, jang2021vowel}. For computer vision tasks, it would also be interesting to derive update rules for convolutional layers.
Another interesting route to explore is the stability of the training process. Although the global convergence of the method cannot be ensured, damping and batching are generally sufficient for the algorithm to converge. However, the impact of the additional constraints of the local divergence minimization problems and the initialization of the approximating distributions are worthy of deeper analysis.



\appendices

\ifCLASSOPTIONcaptionsoff
  \newpage
\fi

\bibliographystyle{IEEEtran}
\bibliography{biblio.bib}

\end{document}